\documentclass{article}

\usepackage[main,final]{neurips_2026}
\usepackage[absolute,overlay]{textpos}
\usepackage{tikz}
\usepackage{hyperref}
\usepackage{url}
\usepackage{amsthm}
\usepackage[table]{xcolor}
\usepackage{hyperref}
\usepackage[most]{tcolorbox}
\usepackage{fontawesome5}
\usepackage[smartEllipses]{markdown}

\usepackage{comment}
\usepackage[most]{tcolorbox}
\usepackage{wrapfig}
\usepackage{multirow}
\usepackage{graphicx,xcolor,float}
\usepackage{threeparttable}
\usepackage[ruled,vlined]{algorithm2e}
\usepackage{colortbl}
\usepackage{color}
\usepackage{multirow}
\usepackage{tabularx}
\usepackage{float}
\usepackage{amsmath, amssymb}
\usepackage{tikz}
\usetikzlibrary{shapes.geometric, arrows.meta, positioning, calc, fit, backgrounds}
\usepackage{graphicx}
\usepackage{booktabs}
\usepackage{arydshln}
\usepackage{enumitem}
\usepackage{wrapfig}
\usepackage{caption}
\usepackage{graphicx}
\usepackage{makecell}
\usepackage{float} 
\usepackage{soul}
\usepackage{mathtools}
\usepackage{bbding}
\usepackage{makecell}
\usepackage{tabularx}
\usepackage{amssymb,mathrsfs,amsmath}
\usepackage{pifont}
\usepackage{bbm}
\usepackage{multirow}   
\usepackage{subcaption}
\usepackage{wrapfig}

\usepackage{xspace}

\newcommand{\Attn}{\operatorname{Attn}}
\newcommand{\Update}{\operatorname{Update}}

\newcommand{\normop}{\operatorname{norm}}

\usepackage{fontawesome5}

\usepackage{listings}

\definecolor{avgpink}{rgb}{1.0, 0.92, 0.92}
\definecolor{bittersweet}{rgb}{1.0, 0.44, 0.37}
\definecolor{mygreen}{rgb}{0.29, 0.7, 0.48}
\usepackage{pifont}
\definecolor{demphcolor}{RGB}{144,144,144}

\definecolor{mygray}{gray}{0.4}
\definecolor{avggray}{gray}{0.9}
\definecolor{autopurple}{HTML}{7030A0}
\definecolor{dyna_yellow}{HTML}{BF9000}
\definecolor{adaptive_blue}{HTML}{0070C0}
\definecolor{darksalmon}{rgb}{0.91, 0.59, 0.48}
\definecolor{emerald}{rgb}{0.31, 0.78, 0.47}
\definecolor{green(pigment)}{rgb}{0.0, 0.65, 0.31}
\definecolor{amaranth}{rgb}{0.9, 0.17, 0.31}
\definecolor{iris}{rgb}{0.35, 0.31, 0.81}
\definecolor{uu}{rgb}{0.95, 0.51, 0.51}
\definecolor{spirodiscoball}{rgb}{0.06, 0.75, 0.99}
\usepackage{svg}

\definecolor{mygrey}{gray}{0.4}

\definecolor{QuestionColor}{rgb}{0.7, 0.1, 0.1} 
\definecolor{AnswerColor}{rgb}{0.1, 0.5, 0.1} 
\definecolor{ReasoningColor}{rgb}{0.1, 0.1, 0.7} 

\usepackage[table,xcdraw,dvipsnames]{xcolor}
\usepackage{cleveref}

\usepackage[utf8]{inputenc} 
\usepackage[T1]{fontenc}    
\usepackage{hyperref}       
\usepackage{url}            
\usepackage{booktabs}       
\usepackage{amsfonts}       
\usepackage{nicefrac}       
\usepackage{microtype}      
\usepackage{xcolor}         

\usepackage{CJKutf8}

\newcommand{\deltamem}{$\delta$-mem}
\title{\deltamem{}: Efficient Online Memory For Large Language Models}

\author{
  Jingdi Lei$^{1\dag}$\thanks{Equal Contribution},\enspace 
  Di Zhang$^{2\dag*}$,\enspace 
  Junxian Li$^{3}$,\enspace 
  Weida Wang$^{2}$,\enspace 
  \textbf{Kaixuan Fan}$^{4}$,\enspace \\ 
  \textbf{Xiang Liu}$^{5}$,\enspace 
  \textbf{Qihan liu}$^{6}$,\enspace 
  \textbf{Xiaoteng Ma}$^{6}$,\enspace 
  \textbf{Baian Chen}$^{6}$,\enspace 
  \textbf{Soujanya Poria}$^{1}$\thanks{Corresponding author} \enspace
  \\
  \textsuperscript{1}\small{Nanyang Technological University},\enspace
  \textsuperscript{2}\small{Fudan University}\\
  \textsuperscript{3}\small{Shanghai Jiao Tong University} 
  \textsuperscript{4}\small{The Chinese University of Hong Kong} \\
  \textsuperscript{5}\small{The Hong Kong University of Science and Technology (Guangzhou)}
  \textsuperscript{6}\small{Mind Lab} \\
  \small{\texttt{jingdi001@e.ntu.edu.sg}}, \ \small{\texttt{di.zhang@ustc.edu}}, \\
  \small{\texttt{soujanya.poria@ntu.edu.sg}}
  \vspace{-5.mm}
}

\begin{document}

\maketitle

\begin{abstract}
Large language models increasingly need to accumulate and reuse historical information in long-term assistants and agent systems. Simply expanding the context window is costly and often fails to ensure effective context utilization. We propose $\delta$-mem, a lightweight memory mechanism that augments a frozen full-attention backbone with a compact online state of associative memory. $\delta$-mem compresses past information into a fixed-size state matrix updated by delta-rule learning, and uses its readout to generate low-rank corrections to the backbone's attention computation during generation. With only an $8\times8$ online memory state, $\delta$-mem improves the average score to $1.10\times$ that of the frozen backbone and $1.15\times$ that of the strongest non-$\delta$-mem memory baseline. It achieves larger gains on memory-heavy benchmarks, reaching $1.31\times$ on \textsc{MemoryAgentBench} and $1.20\times$ on \textsc{LoCoMo}, while largely preserving general capabilities. These results show that effective memory can be realized through a compact online state directly coupled with attention computation, without full fine-tuning, backbone replacement, or explicit context extension. Code is available at:
\href{https://github.com/declare-lab/delta-Mem}{https://github.com/declare-lab/delta-Mem}.
\end{abstract}

\section{Introduction}
\label{sec:intro}

As large language models (LLMs) are increasingly deployed in memory-heavy scenarios requiring continuous interaction, such as long-term personalized assistants~\citep{packer2023memgpt, jiang2025personamem} and long-horizon agent systems~\citep{yao2022react, openai2026codex, anthropic2026claude_code}, their life-cycle must go beyond responding to isolated prompts and instead accumulate, update, and reuse historical information over extended memory-heavy tasks~\citep{yao2022react,shinn2023reflexion,packer2023memgpt,wang2025mirix,zhang2025agent}. In these settings, model performance depends not only on understanding the current input, but also on effectively leveraging relevant past context during test-time~\citep{laban2025llms, zhong2024memorybank}.
An intuitive way is to simply expand the input context and retain more interaction history. However, this strategy only reduces the memory problem to a long-context processing problem, which is both computationally expensive and increasingly difficult to harness. On the one hand, standard attention incurs quadratic cost with respect to context length~\citep{yuan2025native, lei2025error, team2025kimi}. On the other hand, simply increasing the context window does not guarantee effective use of the additional information, as models often suffer from context degradation or context rot when the context becomes very long~\citep{hong2025contextrot, du2025context}, which suggests that even million-token context windows~\citep{openai2026gpt55, google2025gemini3} do not fundamentally solve the memory problem. These limitations call for more advanced memory mechanisms (MMs) that can represent historical information more compactly within a given context window, maintain it dynamically across interactions, and make it effectively usable by the backbone model during test-time~\citep{zhong2024memorybank,packer2023memgpt,wang2023augmenting,wang2025mirix,behrouz2024titans,wang2025m+,zhang2025agent}.

From a unified perspective, existing memory mechanisms can be characterized along two dimensions under a given context window: memory state, which defines how historical information is stored, and memory steering, which determines how stored information influences backbone reasoning. Under this framework, prior methods fall into three paradigms.
Textual memory mechanisms (TMMs)~\citep{packer2023memgpt,zhong2024memorybank,pan2024llmlingua,borgeaud2022improving,chhikara2025mem0} store memory as text and inject it through the input context, offering flexibility without architectural changes but suffering from context-window limits, retrieval noise, and inevitable compaction loss. Outside-channel memory mechanisms (OMMs)~\citep{wu2022memorizing,wang2023augmenting,wei2026mlpmemory} keep memory in external modules and interact with the backbone via retrieval or encoding on outside pathways, enabling modularity but introducing overhead, integration complexity, and potential misalignments with the backbone. Parametric memory mechanisms (PMMs)~\citep{hu2022lora,li2021prefix,meng2022locating,meng2022mass} encode memory into parameters of prefixes or adapters, making them efficient and compatible with frozen backbones, but their static nature limits adaptation to dynamically evolving information. Taken together, these limitations point to a need for a memory mechanism that can maintain a compact and dynamically evolving memory state while steering the backbone through a pathway tightly aligned with its internal attention computation.

Following this motivation, we propose \deltamem{}, a memory mechanism that keeps a compact and dynamically updated memory alongside a frozen full-attention backbone. Instead of storing all historical tokens in the input context, \deltamem{} compresses past information into an online state of associative memory~(OSAM). This state is continuously updated via delta-rule learning as new tokens arrive, allowing the model to maintain useful historical information in a fixed-size matrix representation of associative memories.
During generation, \deltamem{} does not simply retrieve text from memory. Instead, the current input queries the online state to extract context-relevant associative memory signals, which are then transformed into a low-rank correction to the backbone’s attention components. In this way, associative memory directly participates in the backbone’s forward computation while leaving the backbone frozen. The online state is further updated after each interaction, enabling \deltamem{} to evolve its associative memory over time.

Finally, we evaluate \deltamem{} on memory-heavy benchmarks, including \textsc{HotpotQA}~\citep{yang2018hotpotqa}, \textsc{LoCoMo}~\citep{maharana2024evaluating}, and \textsc{MemoryAgentBench}~\citep{hu2025evaluating}, together with general capability benchmarks \textsc{IFEval}~\citep{zhou2023instructionfollowingevaluationlargelanguage} and \textsc{GPQA-Diamond}~\citep{rein2023gpqa}. With only a fixed $8\times8$ online state of associative memory, \deltamem{} improves the final average score by $1.10\times$ over the frozen backbone and outperforms the strongest non-\deltamem{} memory baseline by $1.15\times$. On memory-heavy tasks, the improvement is larger: MemoryAgentBench increases over $1.31\times$, LoCoMo over $1.20\times$, and the TTL subtask nearly doubles from $26.14$ to $50.50$. These results show that a compact online state, when directly coupled with attention computation, can provide effective associative memory without relying on extending explicit context or heavy external retrieval modules. 

Our contributions can be summarized as follows:

\begin{itemize}

    \item We propose \textbf{\deltamem{}}, a memory mechanism that augments a frozen full-attention backbone with a compact online state of associative memory, enabling historical information to be dynamically maintained and directly coupled with the backbone's attention computation.

    \item We show that an extremely small memory state, implemented as an $8 \times 8$ matrix, can retain useful historical signals through OSAM and help the model recover context-relevant information even after explicit history is removed.

    \item We evaluate \deltamem{} on multiple memory-heavy and general capability benchmarks with significant gains on memory-heavy tasks such as \textsc{MemoryAgentBench} and \textsc{LoCoMo}, without full fine-tuning or replacing the backbone architecture.
\end{itemize}

\vspace{-1em}
\section{Preliminaries}
\label{sec:preliminary}
In terms of a Transformer for sequence modeling, let $\mathbf x\in\mathbb{R}^{N\times d}$ denote the input hidden sequence of a selected Transformer layer, where $N$ is the sequence length and $d$ is the hidden dimension. The hidden state at a single position is denoted as $\mathbf x_t\in\mathbb{R}^{d}$. For concise notation, each single-position vector is treated as a column vector. The sequence form can be understood as stacking these vectors along the position dimension. We use $\mathbf Q,\mathbf K,\mathbf V$ to denote the query, key, and value in attention, and use $\mathbf S_t$ to denote the online state after processing position $t$. Unless otherwise specified, we omit the layer index in the following.

Concretely, \deltamem{} maintains a matrix $\mathbf{S}$ as the online state of associative memory. As tokens are processed, this state is updated sequentially to compactly encode key--value associations from the historical context. Given a memory key $\mathbf k_t\in\mathbb{R}^{r}$ and value $\mathbf v_t\in\mathbb{R}^{r}$ at position $t$, the state is expected to store the association $\mathbf k_t\mapsto \mathbf v_t$. The prediction made by the previous state is
\begin{equation}
\hat{\mathbf v}_t
=
\mathbf S_{t-1}\mathbf k_t .
\end{equation}
This memory update can then be regarded as optimizing an online regression loss using SGD:
\begin{equation}
\mathcal{L}_t(\mathbf{S}) = \frac{1}{2}\left\|\mathbf{S}\mathbf{k}_t - \mathbf{v}_t\right\|^2,\quad
\mathbf{S}_t
= \mathbf{S}_{t-1} - \beta_t \nabla_{\mathbf{S}_{t-1}} \mathcal{L}_t(\mathbf{S}_{t-1})
= \mathbf{S}_{t-1} + \beta_t
\left(\mathbf{v}_t-\mathbf{S}_{t-1}\mathbf{k}_t\right)\mathbf{k}_t^{\top},
\end{equation}

This formulation writes only the residual information along the key direction. Consequently, well-learned associations induce negligible updates, whereas predictive discrepancies dynamically correct the memory state. Inspired by gated retention design in Qwen-Next~\citep{qwen3}, we further introduce a forget gate to control long-range state evolution:
\begin{equation}
\mathbf S_t
=
\lambda_t\mathbf S_{t-1}
+
\beta_t
(\mathbf v_t-\mathbf S_{t-1}\mathbf k_t)
\mathbf k_t^\top .
\end{equation}
Here $\lambda_t$ controls how much previous memory is retained, while $\beta_t$ controls the strength of the residual write. This gated delta update forms the basis of the stable online memory dynamics in \deltamem{}.

\section{\deltamem{}}
\label{sec:deltamem}
\begin{figure}[t]
    \centering
    \includegraphics[width=0.95\textwidth]{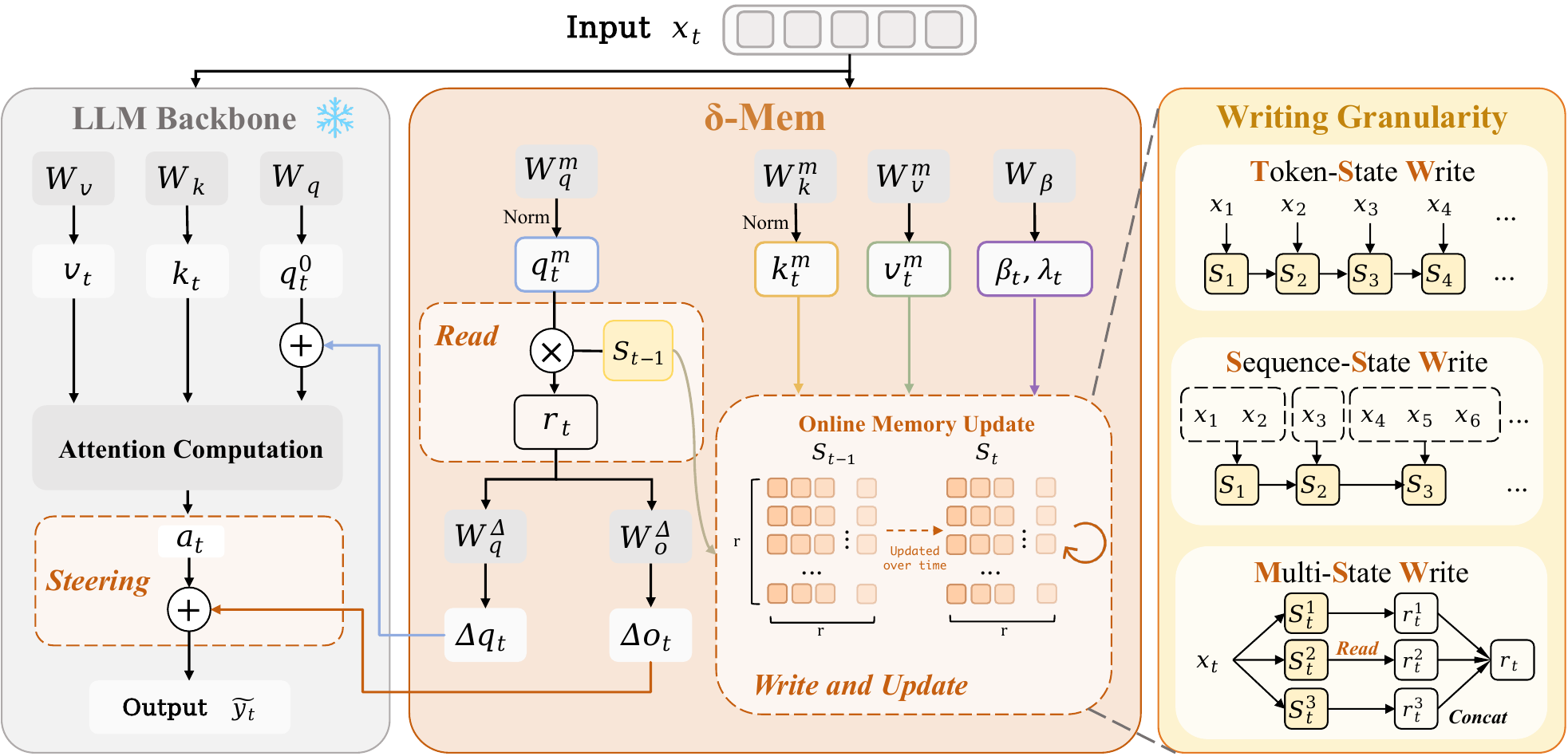}
    \caption{The overview of \deltamem{}. Given the hidden state from a frozen Transformer backbone, \deltamem{} first projects it into a low-dimensional memory space, and reads context-relevant associative memory signals from the previous online state, and uses the signal to generate low-rank corrections to the attention computation. After the computation, the memory state is updated with the current key-value information via delta-rule learning.}
    \label{fig:pipeline}
\end{figure}

At each position, \deltamem{} follows the same computation order: read associative memory signals from the old state, use the signals to steer attention, and then write the current information into the state. In this way, the model can compress history into a state that evolves with the sequence and use it in later reasoning, without updating the backbone parameters. Figure~\ref{fig:pipeline} provides an overview of this design. The frozen backbone performs the standard attention computation, while \deltamem{} reads from the previous state, generates query-side and output-side attention corrections, and updates the online state with the current memory key-value information. The figure also summarizes the three writing strategies studied in this work, corresponding to token-level updates, segment-level updates, and multi-state memory organization.

\subsection{Memory Projections}

To form the online state of associative memory, given a hidden state $\mathbf x_t\in\mathbb{R}^{d}$ at the current position, \deltamem{} projects it into a low-dimensional associative memory space:
\begin{equation}
\mathbf q_t^m
=
L_2\normop
\left(
\tanh(\mathbf W_q^m \mathbf x_t)
\right), \quad
\mathbf k_t^m
=
L_2\normop
\left(
\tanh(\mathbf W_k^m \mathbf x_t)
\right), \quad
\mathbf v_t^m
=
\mathbf W_v^m \mathbf x_t,
\end{equation}
where $\mathbf q_t^m,\mathbf k_t^m,\mathbf v_t^m\in\mathbb{R}^{r}$. These three vectors correspond to reading and writing in memory. $\mathbf q_t^m$ queries the old state, while $\mathbf k_t^m$ and $\mathbf v_t^m$ describe how the current information should be written into the state. Normalizing the query and key can reduce state instability caused by scale drift during long-sequence recurrence.

The write gate and retention gate are also determined by the current hidden state:
\begin{equation}
\boldsymbol\beta_t=\sigma(\mathbf W_\beta \mathbf x_t+\mathbf b), \quad
\boldsymbol\lambda_t=\mathbf 1-\boldsymbol\beta_t.
\end{equation}
where $\boldsymbol\beta_t,\boldsymbol\lambda_t\in\mathbb{R}^{r}$, $\mathbf b$ is the bias, and $\sigma$ is the sigmoid function. This allows the state update to be adjusted dimension by dimension: some dimensions retain old memory, while others write the current information more actively.

\subsection{Reading from Online State of Associative Memory}

Before writing the current information, \deltamem{} first reads from the old state:
\begin{equation}
\mathbf r_t
=
\mathbf S_{t-1}\mathbf q_t^m.
\end{equation}
The read vector $\mathbf r_t\in\mathbb{R}^{r}$ is the result of querying the online memory state with the current input. Since the size of $\mathbf S_{t-1}$ is fixed, the cost of this step is independent of the history length.

This reading form is complementary to standard attention. Attention compares the query with all keys within the explicit context, while \deltamem{} directly obtains continuous associative memory signals from the compressed state. It does not return text segments or add context tokens. Instead, it provides history-dependent steering signals before the attention computation.

\subsection{Steering Attention through Low-Rank Corrections}

The associative memory signals steer the attention computation through two lightweight linear mappings. 
First, the read signal $\mathbf r_t$ is projected into a query-side correction and an output-side correction:
\begin{equation}
\Delta \mathbf q_t = \mathbf W_q^\Delta \mathbf r_t,
\qquad
\Delta \mathbf o_t = \mathbf W_o^\Delta \mathbf r_t .
\end{equation}

The query-side correction is then added to the original query of the frozen backbone:
\begin{equation}
\mathbf q_t^0 = \mathbf W_Q \mathbf x_t,
\qquad
\tilde{\mathbf q}_t =
\mathbf q_t^0+\frac{\alpha}{r}\Delta \mathbf q_t .
\end{equation}

The attention output $\mathbf a_t$ is then computed using the corrected query and the frozen backbone keys and values, while the output-side correction is added after attention:
\begin{equation}
\mathbf a_t =
\Attn(\tilde{\mathbf q}_t,\mathbf K_{\leq t},\mathbf V_{\leq t}),
\qquad
\tilde{\mathbf y}_t =
\mathbf a_t+\frac{\alpha}{r}\Delta \mathbf o_t .
\end{equation}

The main implementation only uses the two correction terms on the query and output sides, and we detail these choices in Section~\ref{sec:ablations}. The low-rank correction here is different from a static adapter. Although $\mathbf W_q^\Delta$ and $\mathbf W_o^\Delta$ are fixed after training, their input $\mathbf r_t$ comes from the dynamic state $\mathbf S_{t-1}$. Therefore, the same set of parameters can produce different steering effects under different histories.

\subsection{Writing into Online State of Associative Memory}

After the current attention computation is completed, \deltamem{} writes the information at the current position into the online state. Given the current memory key-value pair $(\mathbf k_t^m,\mathbf v_t^m)$, the previous state first predicts the value associated with the current key as $\mathbf S_{t-1}\mathbf k_t^m$. The difference between the target value and this prediction defines the residual information to be written. As described in Section~\ref{sec:preliminary} \deltamem{} updates the state with a dimension-wise gated delta-rule:
\begin{equation}
\mathbf S_t
=
\mathrm{Diag}(\boldsymbol\lambda_t)\mathbf S_{t-1}
+
\mathrm{Diag}(\boldsymbol\beta_t)
\left(
\mathbf v_t^m-\mathbf S_{t-1}\mathbf k_t^m
\right)
(\mathbf k_t^m)^\top .
\end{equation}

Expanding the update gives:
\begin{equation}
\mathbf S_t
=
\mathrm{Diag}(\boldsymbol\lambda_t)\mathbf S_{t-1}
-
\mathrm{Diag}(\boldsymbol\beta_t)
\mathbf S_{t-1}
\mathbf k_t^m(\mathbf k_t^m)^\top
+
\mathrm{Diag}(\boldsymbol\beta_t)
\mathbf v_t^m(\mathbf k_t^m)^\top .
\end{equation}

The three terms have clear roles: the first term retains the previous state, the second term removes the old prediction component along the current key direction, and the third term writes the new value into the same direction. Thus, the memory state is updated by error correction with controlled forgetting, rather than by unselectively accumulating new outer products.

The dimension-wise nature of the gates can be seen by expanding the update row by row. Let $\mathbf s_t^{(i)}$ denote the $i$-th row of $\mathbf S_t$. Then,
\begin{equation}
\mathbf s_t^{(i)}
=
\lambda_{t,i}\mathbf s_{t-1}^{(i)}
+
\beta_{t,i}
\left(
v_{t,i}^m-\mathbf s_{t-1}^{(i)}\mathbf k_t^m
\right)
(\mathbf k_t^m)^\top .
\end{equation}
This shows that each memory dimension can independently control how much old information is retained and how strongly the current residual is written. Such dimension-wise gating is useful for continuous interactions, where the state must preserve stable historical information while still adapting to new inputs.

\subsection{Writing Granularity of Online State}

The above formulas explain how a single write operation is performed, but the memory mechanism also depends on the definition of writing granularity. A token is the finest granularity, but it is not always the most suitable one. In conversations and agent trajectories, messages, semantic segments, or stage-level events are often more stable. We therefore examine three writing strategies. As illustrated in Figure~\ref{fig:pipeline}, TSW writes at every token, SSW averages the hidden states within each segment and writes per segment, and MSW writes into multiple parallel sub-states and then aggregates their readouts.

\paragraph{Token-State Write~(TSW).}
Token-State Write updates the online state at each token position:
\begin{equation}
\mathbf S_t
=
\Update(\mathbf S_{t-1},\mathbf x_t).
\end{equation}
It preserves the finest-grained information and is suitable for scenarios that need to capture local changes. However, since every token triggers a write operation, the state is also more easily affected by format symbols, repeated expressions, and short-term noise.

\paragraph{Sequence-State Write~(SSW).}
Sequence-State Write raises the writing granularity from unit tokens to a message segment. Let $\mathcal{M}^{(j)}$ denote the set of token indices in the $j$-th message. We first obtain the segment representation by averaging the hidden states of all tokens within this message:
\begin{equation}
\bar{\mathbf x}^{(j)} = \frac{1}{|\mathcal{M}^{(j)}|} \sum_{t \in \mathcal{M}^{(j)}} \mathbf x_t.
\end{equation}

Then, each message updates the online state once. Let $\mathbf S_{(j)}$ denote the state after incorporating the $j$-th message:
\begin{equation}
\mathbf S_{(j)} = \Update(\mathbf S_{(j-1)},\bar{\mathbf x}^{(j)}).
\end{equation}
SSW reduces redundant writes and smooths the state evolution. The cost is that some fine-grained token-level details are absorbed by the averaged segment representation.

\paragraph{Multi-State Write~(MSW).}
The first two strategies adjust the writing granularity, while MSW adjusts the state organization. A single state needs to contain facts, preferences, task progress, and local events at the same time, which may easily lead to overwriting and interference. MSW decomposes memory into multiple parallel sub-states:
\begin{equation}
\mathcal S_t=\{\mathbf S_t^{(1)},\ldots,\mathbf S_t^{(N)}\},\qquad
\mathbf S_t^{(i)}=\Update^{(i)}(\mathbf S_{t-1}^{(i)},\mathbf x_t),\qquad
\mathbf r_t=Concat(\mathbf r_t^{(1)},\ldots,\mathbf r_t^{(N)}).
\end{equation}
where $N$ is the number of state, $\mathbf S_t^{(i)}\in\mathbb{R}^{r\times r}$, and $\mathbf r_t^{(i)}=\mathbf S_{t-1}^{(i)}\mathbf q_t^{m,(i)}$ for $i=1,\ldots,N$. This organization allows different sub-states to accumulate different types of information, thereby reducing mutual interference within a single state.

\subsection{Training Objective}
\deltamem{} is trained with the standard SFT loss. For each example, the context tokens are first written into the online state, producing $\mathbf S_C$, while they are not replayed as explicit backbone input during prediction. The frozen backbone only receives the query $Q$ and response $Y$, and the stored state steers attention through \deltamem{}. The loss is the autoregressive cross-entropy over response tokens:
\begin{equation}
\mathcal L_{\mathrm{SFT}}
=
-\sum_{j=1}^{|Y|}
\log p_{\phi,\theta}(y_j \mid Q, y_{<j}, \mathbf S_C),
\end{equation}
where $\theta$ denotes the trainable \deltamem{} parameters and $\phi$ denotes the frozen backbone parameters.

\section{Experiments}
\label{sec:experiments}
\subsection{Experimental Setup}
\vspace{-0.5em}

\paragraph{Evaluation and Benchmarks.} To independently measure general reasoning and memory effectiveness, we evaluate our method on general tasks and memory-heavy benchmarks. General multi-hop reasoning, knowledge-intensive QA, and instruction-following are assessed using HotpotQA~\citep{yang2018hotpotqa}, GPQA-Diamond~\citep{rein2023gpqa}, and IFEval~\citep{zhou2023instructionfollowingevaluationlargelanguage}. For the memory-heavy side, we utilize LoCoMo~\citep{maharana2024evaluating} (following~\citep{chhikara2025mem0}, the adversarial question category is excluded), alongside MemoryAgentBench~\citep{hu2025evaluating} to evaluate the retention, retrieval, and utilization of memory information across extended interaction histories. 

\vspace{-0.5em}
\paragraph{Baselines.}
We compare \deltamem{} against representative memory mechanisms. All methods are built on the same Qwen3-4B-Instruct backbone. For textual memory mechanisms, we consider BM25 RAG~\citep{lewis2020retrieval}, which retrieves relevant historical text and prepends it to the context; LLMLingua-2~\citep{pan2024llmlingua}, which compresses long histories into a shorter textual context; and MemoryBank~\citep{zhong2024memorybank}, which maintains continuous interaction history through textual memory entries. For parametric memory mechanisms, we compare with Context2LoRA~\citep{hu2022lora,back2026understanding} and MemGen~\citep{zhang2025memgen}, which encode memory or context-dependent adaptation into additional trainable parameters. For outside-channel memory, we include an MLP Memory~\citep{wei2026mlpmemory} baseline that retrieves information in a separate module and then fuses back into the model. We additionally report trainable parameter counts for rank-8 configurations to compare memory effectiveness under similar or smaller adaptation budgets in Appendix~\ref{appendix:parameter}.

\vspace{-0.5em}
\paragraph{Implementation Details}
We select LLM backbones of varying sizes, including Qwen3-8B~\citep{qwen3}, Qwen3-4B-Instruct~\citep{qwen3}, and SmolLM3-3B~\citep{bakouch2025smollm3}. More
training setup and evaluation configurations are listed in Appendix~\ref{appendix:setup}.

\subsection{Main Results across Memory Mechanisms}
\vspace{-1em}
\begin{table}[htbp]
    \centering
    \caption{Main benchmark results comparing different memory mechanisms on Qwen3-4B-Instruct. All values represent the performance metric for each task is detailed in Appendix~\ref{appendix:setup}. For the final average score, HotpotQA is counted using Exact Match~(EM).}
    \label{tab:baseline}
    \resizebox{\textwidth}{!}{
    \begin{tabular}{lccccccccccccccc}
    \toprule
    \multirow{2}{*}{\textbf{Model}} & \multirow{2}{*}{\textbf{IFEval}} & \multicolumn{2}{c}{\textbf{HotpotQA}} & \multirow{2}{*}{\textbf{GPQA-D}} & \multicolumn{5}{c}{\textbf{Memory Agent Bench}} & \multicolumn{5}{c}{\textbf{LoCoMo}} & \multirow{2}{*}{\textbf{Avg.}} \\
    \cmidrule(lr){3-4} \cmidrule(lr){6-10} \cmidrule(lr){11-15}
    & & \textbf{EM} & \textbf{F1} & & \textbf{Avg.} & \textbf{AR} & \textbf{TTL} & \textbf{LRU} & \textbf{SF} & \textbf{Avg.} & \textbf{Multi} & \textbf{Temp} & \textbf{Open} & \textbf{Single} \\
    \midrule
    \textbf{Qwen3-4B-Instruct} & 81.89 & 42.35 & 56.00 & 39.39 & 29.54 & 35.30 & 26.14 & \textbf{47.08} & 14.37 & 40.79 & 38.39 & 32.89 & 10.77 & 48.05 & 46.79 \\ 
    \rowcolor{gray!15}
    \multicolumn{16}{c}{\textbf{Textual Memory}} \\
   \quad + BM25 RAG & - & 40.35 & 52.83 & - & 24.49 & 32.20 & 9.74 & 37.86 & 15.00 & 36.68 & 38.12 & 20.34 & 9.99 & 45.47 & 44.56 \\
   \quad + LLMLingua-2 & - & 36.93 & 50.03 & - & 15.63 & 21.45 & 1.43 & 38.45 & 8.62 & 40.98 & 39.07 & 30.13 & 10.98 & 49.19 & 42.96 \\
    \quad + MemoryBank  & - & - & - & - & 17.65 & 22.65 & 7.67 & 36.36 & 9.88 & 38.14 & 37.88 & 21.76 & 13.35 & 47.31 & 43.88 \\ 
    \rowcolor{gray!15}
    \multicolumn{16}{c}{\textbf{Parametric Memory}} \\
    \quad + Context2LoRA & 76.71 & 37.85 & 50.88 & 29.29 & 32.53 & 40.00 & 29.86 & 25.15 & \textbf{17.75} & 48.11 & 37.95 & 34.99 & 16.75 & \textbf{60.11} & 44.90 \\
    \quad + MemGen   & 39.37 & 5.36 & 16.27 & 38.89 & 29.61 & 34.85 & 28.45 & 44.30 & 14.38 & 40.05 & 32.93 & 33.30 & 12.67 & 48.13 & 30.66 \\
    \rowcolor{gray!15}
    \multicolumn{16}{c}{\textbf{Outside-channel Memory}} \\
    \quad + MLP Memory  & 24.95 & 10.94 & 25.83 & 22.73 & 28.80 & 35.35 & 26.00 & 31.19 & 14.38 & 26.85 & 32.87 & 16.72 & 8.81 & 30.75 & 22.85 \\
    \midrule
    \rowcolor{gray!15}
    \multicolumn{16}{c}{$\delta$\textbf{-Mem}} \\
    \quad + $\delta$-Mem~(SSW) & 81.70 & 49.22 & 63.43 & \textbf{41.41} & 37.84 & 41.50 & \textbf{50.50} & 43.02 & 16.50 & 47.05 & 41.00 & 36.48 & 14.08 & 56.88 & 51.44 \\
    \quad + $\delta$-Mem~(TSW) & \textbf{82.99} & \textbf{49.41} & \textbf{63.66} & 40.40 & 36.48 & 42.45 & 40.64 & 46.08 & 15.88 & 46.53 & 42.14 & 37.20 & 13.35 & 55.36 & \textbf{51.66} \\
    \quad + $\delta$-Mem~(MSW) & 81.52 & 46.86 & 60.47 & 37.37 & \textbf{38.85} & \textbf{44.40} & 47.29 & 41.55 & 17.00 & \textbf{49.12} & \textbf{42.57} & \textbf{39.31} & \textbf{18.12} & 58.59 & 50.74 \\
    \bottomrule
    \end{tabular}
    }
\end{table}

Table~\ref{tab:baseline} compares \deltamem{} with representative memory-augmented baselines on general reasoning, instruction following, and memory-heavy benchmarks. \deltamem{} achieves the strongest performance across all methods. The TSW variant reaches the best average score of 51.66\%, improving over the Qwen3-4B-Instruct backbone (46.79\%) by +4.87 points and over Context2LoRA (44.90\%) by +6.76 points. SSW and MSW also perform strongly, achieving 51.44\% and 50.74\%, respectively. The gains are most pronounced on memory-heavy benchmarks. On MemoryAgentBench, \deltamem{} improves the average score from 29.54\% to 38.85\%, with MSW performing best. On LoCoMo, MSW achieves the highest average of 49.12\% and performs best on Multi, Temporal, and Open subsets. On HotpotQA, TSW improves EM/F1 from 42.35\%/56.00\% to 49.41\%/63.66\%.

Across baselines, different memory mechanisms exhibit distinct limitations. Textual memory methods show inconsistent gains, likely due to retrieval noise and information loss introduced by compressing memory into token space. Parametric memory methods such as Context2LoRA tend to generalize less robustly across tasks, as their memory is statically encoded in parameters and can overfit to training distributions. The MLP Memory baseline performs relatively limited, indicating it lacks sequential state accumulation and cannot explicitly model long-range dependencies, while also introducing information loss by approximating instance-level retrieval. In contrast, \deltamem{} consistently improves performance across both general and memory-heavy evaluations, suggesting that maintaining memory as an online state provides a more robust memory mechanism.

\subsection{Main Results Across Different Backbone Models}

\begin{table*}[htbp]
    \centering
    \caption{General benchmark and long-context evaluation results across backbone models. All values represent the performance metric for each task as detailed in Appendix~\ref{appendix:setup}. For the final average score, HotpotQA is counted using Exact Match~(EM).}
    \label{tab:general_results}
    \resizebox{\textwidth}{!}{
    \begin{tabular}{lccccccccccccccc}
    \toprule
    \multirow{2}{*}{\textbf{Model}} & \multirow{2}{*}{\textbf{IFEval}} & \multicolumn{2}{c}{\textbf{HotpotQA}} & \multirow{2}{*}{\textbf{GPQA-D}} & \multicolumn{5}{c}{\textbf{Memory Agent Bench}} & \multicolumn{5}{c}{\textbf{LoCoMo}} & \multirow{2}{*}{\textbf{Avg.}} \\
    \cmidrule(lr){3-4} \cmidrule(lr){6-10} \cmidrule(lr){11-15}
    & & \textbf{EM} & \textbf{F1} & & \textbf{Avg.} & \textbf{AR} & \textbf{TTL} & \textbf{LRU} & \textbf{SF} & \textbf{Avg.} & \textbf{Multi} & \textbf{Temp} & \textbf{Open} & \textbf{Single} \\
    \midrule
    \textbf{Qwen3-4B-Instruct} & 81.89 & 42.35 & 56.00 & 39.39 & 29.54 & 35.30 & 26.14 & \textbf{47.08} & 14.37 & 40.79 & 38.39 & 32.89 & 10.77 & 48.05 & 46.79 \\
    \quad + $\delta$-Mem (SSW) & 81.70 & 49.22 & 63.43 & \textbf{41.41} & 37.84 & 41.50 & \textbf{50.50} & 43.02 & 16.50 & 47.05 & 41.00 & 36.48 & 14.08 & 56.88 & 51.44 \\
    \quad + $\delta$-Mem (TSW) & \textbf{82.99} & \textbf{49.41} & \textbf{63.66} & 40.40 & 36.48 & 42.45 & 40.64 & 46.08 & 15.88 & 46.53 & 42.14 & 37.20 & 13.35 & 55.36 & \textbf{51.66} \\
    \quad + $\delta$-Mem (MSW) & 81.52 & 46.86 & 60.47 & 37.37 & \textbf{38.85} & \textbf{44.40} & 47.29 & 41.55 & 17.00 & \textbf{49.12} & \textbf{42.57} & \textbf{39.31} & \textbf{18.12} & 58.59 & 50.74 \\
    \quad + Context2LoRA & 76.71 & 37.85 & 50.88 & 29.29 & 32.53 & 40.00 & 29.86 & 25.15 & \textbf{17.75} & 48.11 & 37.95 & 34.99 & 16.75 & \textbf{60.11} & 44.90 \\
    \midrule
    \textbf{Qwen3-8B} & 79.67 & 32.48 & 41.42 & 44.95 & 31.87 & 45.10 & 12.79 & \textbf{48.22} & 12.00 & 47.02 & 41.28 & 30.43 & 14.20 & 59.03 & 47.20 \\
    \quad + $\delta$-Mem (SSW) & 80.41 & \textbf{43.81} & \textbf{56.61} & 45.45 & \textbf{33.63} & 45.55 & \textbf{19.52} & 46.11 & 13.50 & \textbf{51.01}  & \textbf{46.46} & \textbf{35.36} & 20.05 & 62.05 & \textbf{50.86} \\
    \quad + $\delta$-Mem (TSW) & \textbf{82.81} & 41.97 & 53.61 & 44.95 & 32.97 & \textbf{45.65} & 16.43 & 46.41 & 12.88 & 50.70 & 44.35 & 34.76 & 17.90 & \textbf{62.66} & 50.68 \\
    \quad + $\delta$-Mem (MSW) & 80.96 & 40.15 & 51.34 & \textbf{49.49} & 32.66 & 45.55 & 14.95 & 44.60 & 13.38 & 50.92 & 45.52 & 34.82 & \textbf{21.21} & 62.27 & 50.84 \\
    \quad + Context2LoRA & 77.26 & 36.22 & 49.19 & 38.38 & 30.52 & 43.15 & 10.05 & 34.04 & \textbf{16.13} & 47.20 & 35.18 & 35.05 & 20.49 & 58.92 & 45.92 \\
    \midrule
    \textbf{SmolLM3-3B} & 67.10 & 1.67 & 14.40 & 23.23 & 14.21 & 12.57 & 5.53 & 30.72 & 8.00 & 24.18 & 22.41 & 16.49 & 10.87 & 29.22 & 26.08\\
    \quad + $\delta$-Mem (SSW) & \textbf{70.61} & 27.35 & 43.26 & \textbf{26.77} & 19.22 & 16.80 & 6.50 & 37.84 & 15.75 & 39.39 & 29.12 & 27.09 & \textbf{22.57} & 49.45 & 36.67\\
    \quad + $\delta$-Mem (TSW) & 66.36 & 24.90 & 41.28 & 26.26 & \textbf{20.74} & 17.98 & \textbf{8.71} & \textbf{41.63} & 15.75 & 35.46 & 27.48 & 25.34 & 17.04 & 44.10 & 34.74\\
    \quad + $\delta$-Mem (MSW) & 67.47 & \textbf{31.61} & \textbf{46.77} & 25.76 & 20.54 & \textbf{18.10} & 8.32 & 39.63 & \textbf{16.12} & \textbf{39.41} & \textbf{29.78} & \textbf{27.15} & 19.59 & \textbf{49.59} & \textbf{36.96}\\
    \quad + Context2LoRA & 62.29 & 30.28 & 44.39 & \textbf{26.77} & 17.62 & 16.08 & 2.86 & 36.77 & 14.75 & 37.74 & 26.35 & 26.05 & 15.41 & 48.58 & 34.94\\
    \bottomrule
    \end{tabular}
    }
\end{table*}

Table~\ref{tab:general_results} evaluates \deltamem{} across three backbone models, demonstrating consistent improvements in average scores across the board. \deltamem{} improves the average score on all backbones. Specifically, it boosts Qwen3-4B-Instruct from 46.79\% to 51.66\%, Qwen3-8B from 47.20\% to 50.86\%, and SmolLM3-3B from 26.08\% to 36.96\%. Notably, the effectiveness of the writing strategies varies by model capacity. On the more capable Qwen3-8B, the improvements are more modest but steady, with SSW securing the top average score of 50.86\%. This suggests that for backbones with stronger inherent reasoning, segment-level writing (SSW) smooths state updates and effectively mitigates token-level noise. In contrast, the smaller SmolLM3-3B exhibits a substantial performance leap (from 26.08\% to 36.96\%) driven by MSW, indicating that smaller backbones benefit significantly from separating memory into multiple states to minimize interference.

\section{Ablative Study}
\label{sec:ablations}
\subsection{Context Recovery}

\begin{wrapfigure}{r}{0.6\textwidth}
    \centering
    \includegraphics[width=0.6\textwidth]{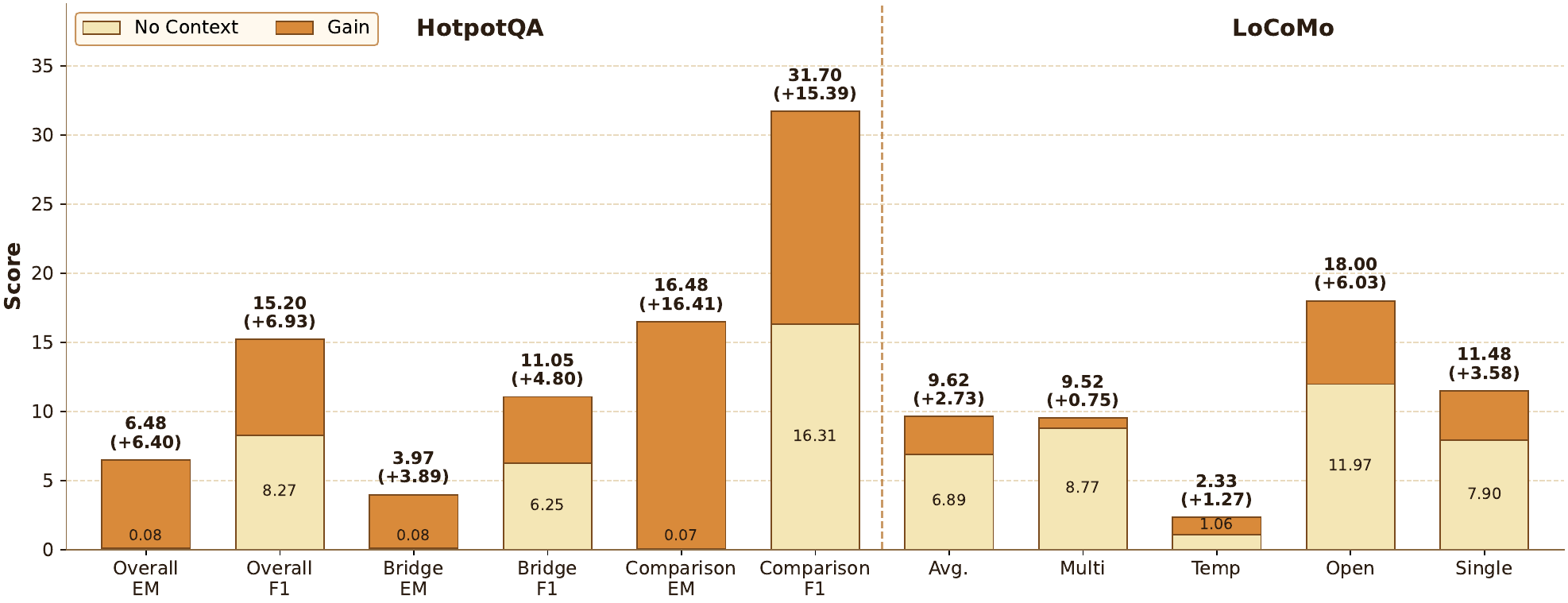}
    \caption{Context recovery performance on HotpotQA and LoCoMo with Qwen3-4B-Instruct as the backbone.}
    \label{fig:context_recover}
    \vspace{-1.0em}
\end{wrapfigure}

To examine whether the online state of associative memory can preserve useful historical information without explicit context replay, we evaluate \deltamem{} under a no-context setting, where the original historical context is removed and only the compressed memory state is injected. As shown in Figure~\ref{fig:context_recover}, \deltamem{} consistently improves over the no-context baseline on both HotpotQA and LoCoMo. On HotpotQA, the overall EM increases from 0.08\% to 6.48\%, and the overall F1 improves from 8.27\% to 15.20\%. The gains are especially large on the Bridge subset, where EM rises from 0.08\% to 3.97\% and F1 increases from 6.25\% to 11.05\%, indicating that the
online state can recover part of the missing multi-hop evidence. On LoCoMo, \deltamem{} also improves the overall average from 3.49\% to 8.05\%, with clear gains across multi-hop, temporal, open-domain, and single-hop questions. These results suggest that the online state of associative memory stores context-relevant historical signals that can be reused when explicit context is unavailable.

\subsection{Heads Ablation}
\vspace{-1em}
\begin{table}[htbp]
\centering
\caption{Head Ablation Results on HotpotQA and LoCoMo with Qwen3-4B-Instruct as the backbone.}
\label{tab:head_ablation}
\resizebox{\textwidth}{!}{
\begin{tabular}{lcccccccccccc}
\toprule
\multirow{3}{*}{\textbf{Head}} & \multicolumn{6}{c}{\textbf{HotpotQA}} & \multicolumn{5}{c}{\textbf{LoCoMo}} & \multirow{3}{*}{\textbf{Avg.}} \\
\cmidrule(lr){2-7} \cmidrule(lr){8-12}
& \multicolumn{2}{c}{\textbf{Overall}} & \multicolumn{2}{c}{\textbf{Bridge}} & \multicolumn{2}{c}{\textbf{Comparison}} & \multirow{2}{*}{\textbf{Avg.}} & \multirow{2}{*}{\textbf{Multi}} & \multirow{2}{*}{\textbf{Temp}} & \multirow{2}{*}{\textbf{Open}} & \multirow{2}{*}{\textbf{Single}} \\
\cmidrule(lr){2-3} \cmidrule(lr){4-5} \cmidrule(lr){6-7}
& \textbf{EM} & \textbf{F1} & \textbf{EM} & \textbf{F1} & \textbf{EM} & \textbf{F1} & & & & & \\
\midrule
q    & 45.87 & 60.59 & 44.96 & 60.34 & 49.50 & 61.58 & 43.15 & \textbf{42.43} & 33.60 & 10.03 & 50.82 & 44.51 \\
k    & 43.39 & 57.28 & 42.35 & 56.82 & 47.55 & 59.09 & 40.98 & 38.44 & 33.12 & 10.93 & 48.26 & 42.19 \\
v    & 46.12 & 60.95 & 45.07 & 60.54 & 50.30 & 62.57 & 42.35 & 39.54 & 34.24 & 11.11 & 49.95 & 44.24 \\
o    & 48.94 & 63.69 & 47.67 & 63.58 & 54.00 & 64.15 & 45.15 & 39.68 & 36.31 & 12.77 & 54.06 & 47.05 \\
\midrule
qk   & 45.46 & 59.62 & 44.19 & 59.12 & 50.50 & 61.59 & 42.94 & 40.64 & 33.65 & 10.09 & 51.01 & 44.20 \\
qv   & 47.02 & 61.95 & 45.86 & 61.64 & 51.65 & 63.20 & 43.24 & 40.38 & 35.57 & 10.76 & 50.84 & 45.13 \\
qo   & 49.41 & 63.66 & 47.65 & 63.22 & \textbf{56.42} & 65.42 & \textbf{46.53} & 42.14 & 37.20 & \textbf{13.35} & \textbf{55.36} & 47.97 \\
kv   & 45.67 & 60.43 & 44.90 & 60.25 & 48.76 & 61.14 & 42.02 & 39.45 & 33.91 & 11.00 & 49.52 & 43.85 \\
\midrule
qko  & 48.24 & 62.42 & 46.79 & 62.11 & 54.00 & 63.66 & 46.01 & 40.89 & 36.82 & 12.61 & 55.05 & 47.13 \\
qkv  & 47.47 & 62.56 & 46.40 & 62.25 & 51.71 & 63.81 & 42.42 & 39.40 & 34.70 & 10.04 & 50.08 & 44.95 \\
\midrule
qkvo & \textbf{49.94} & \textbf{65.01} & \textbf{48.39} & \textbf{64.63} & 56.09 & \textbf{66.56} & 46.15 & 41.08 & \textbf{37.25} & 13.14 & 55.02 & \textbf{48.05} \\
\bottomrule
\end{tabular}
}
\end{table}

We first study where the memory-induced correction should be injected within the attention block. As shown in Table~\ref{tab:head_ablation}, applying \deltamem{} to both query and output branches already yields strong performance, suggesting that query-side and output-side corrections provide an effective interface for memory injection. Among single-branch variants, the output branch performs best, achieving an average score of 47.05\%, while the key branch is less effective. Combining multiple branches further improves performance. The full qkvo configuration achieves the best average score of 48.05\%. These results suggest that associative memory signals are most effective when they can jointly influence query formation, key-value interaction, and output representation. While qkvo yields the highest average score, its marginal gain over qo does not justify the extra parameter overhead. Thus, we default to qo for an optimal performance-efficiency trade-off.

\subsection{Insertion Depth Ablation}
\vspace{-1em}
\begin{table}[htbp]
\centering
\caption{Insertion depth results on HotpotQA and LoCoMo with Qwen3-4B-Instruct as the backbone.}
\label{tab:layer_ablation}
\resizebox{\textwidth}{!}{
\begin{tabular}{lcccccccccccc}
\toprule
\multirow{3}{*}{\textbf{Layer}} & \multicolumn{6}{c}{\textbf{HotpotQA}} & \multicolumn{5}{c}{\textbf{LoCoMo}} & \multirow{3}{*}{\textbf{Avg.}}\\
\cmidrule(lr){2-7} \cmidrule(lr){8-12}
& \multicolumn{2}{c}{\textbf{Overall}} & \multicolumn{2}{c}{\textbf{Bridge}} & \multicolumn{2}{c}{\textbf{Comparison}} & \multirow{2}{*}{\textbf{Avg.}} & \multirow{2}{*}{\textbf{Multi}} & \multirow{2}{*}{\textbf{Temp}} & \multirow{2}{*}{\textbf{Open}} & \multirow{2}{*}{\textbf{Single}} \\
\cmidrule(lr){2-3} \cmidrule(lr){4-5} \cmidrule(lr){6-7}
& \textbf{EM} & \textbf{F1} & \textbf{EM} & \textbf{F1} & \textbf{EM} & \textbf{F1} & & & & & \\
\midrule
Front 12  & 45.52 & 61.01 & 45.32 & 61.08 & 46.33 & 60.77 & 43.26 & 39.06 & 33.65 & 10.19 & 52.10 & 44.39 \\
Middle 12 & 47.44 & 60.59 & 45.56 & 60.06 & 54.94 & 62.71 & 45.87 & 44.00 & 35.75 & 13.09 & 54.10 & 46.66 \\
Back 12   & 44.58 & 59.04 & 43.56 & 58.65 & 48.62 & 60.58 & 43.53 & 40.60 & 36.31 & 11.11 & 50.97 & 44.06 \\
All Layers   & 49.41 & 63.66 & 47.65 & 63.22 & 56.42 & 65.42 & 46.53 & 42.14  & 37.20 & 13.35 & 55.36 & 47.97 \\
\bottomrule
\end{tabular}
}
\end{table}
Table~\ref{tab:layer_ablation} studies the insertion depth of \deltamem{} across model's layers. Applying memory correction to all layers achieves the best overall performance, with an average score of 47.97\%. It also obtains the strongest HotpotQA result, improving the overall EM/F1 to 49.41\%/63.66\%, and reaches the best LoCoMo average of 46.53\%. These results suggest that associative memory signals are most effective when they can influence the representation hierarchy across the full depth of the backbone. Among partial-layer variants, the middle-layer configuration performs best, reaching an average score of 46.66\%. It clearly outperforms both the front-layer and back-layer configurations on the final average score. This indicates that intermediate layers provide a particularly effective interface for memory injection, balancing semantic abstraction and task-specific computation. In contrast, front-layer injections act on overly local representations, while back-layer injections leave insufficient depth for associative memory signals to propagate through subsequent computations.

\section{Related Work}
\label{sec:related_work}

\paragraph{Textual Memory Mechanisms.}
Textual memory mechanisms externalize memory as text entries, summaries, or retrievable documents, and re-inject selected evidence into the input context or retrieval-augmented generation process. Early retrieval-augmented systems~\citep{lewis2020retrieval, borgeaud2022improving} demonstrate the effectiveness of scalable textual stores for knowledge-intensive generation, while later agent-oriented methods~\citep{packer2023memgpt, park2023generative, zhong2024memorybank, chhikara2025mem0} extend this paradigm to continuous interaction by organizing past history and experience through logging, summarization, and reflection. Despite their flexibility, textual memory remains constrained by its tokenized form: memory use is sensitive to compression fidelity, retrieval noise, and context budget~\citep{laban2025llms, hong2025contextrot}. \deltamem{} does not route compressed history back through token space. Instead, it maintains a compact online state and uses its readout to steer the frozen Transformer through low-rank attention corrections, separating memory maintenance from prompt-level reinsertion.

\paragraph{Outside-Channel Memory Mechanisms.}
A related line of work stores memory outside the backbone while preserving it in latent rather than textual form~\citep{wu2022memorizing, wang2023augmenting, wei2026mlpmemory}. Memorizing Transformers~\citep{wu2022memorizing} store past internal representations as non-differentiable key-value memories and retrieve them with approximate kNN, while LongMem~\citep{wang2023augmenting} uses a frozen backbone as a memory encoder and an adaptive residual side network to read from an external memory bank. Compared with textual memory, latent memory can avoid part of the information loss introduced by natural-language summarization and preserve richer internal representations. However, memory still interacts with the backbone through a separate retrieval or reader pathway, introducing retrieval overhead, fusion complexity, and possible mismatch between stored and current representations. \deltamem{} differs in that its memory is not retrieved as an auxiliary external source; instead, its compact online state directly produces low-rank corrections to the attention computation, allowing memory to participate in the current forward pass.

\paragraph{Parametric Memory Mechanisms.}
Parametric memory mechanisms encode memory into additional parameters or localized weight edits. Prefix-Tuning~\citep{li2021prefix} learns continuous virtual tokens for a frozen model, while LoRA~\citep{hu2022lora} injects low-rank trainable updates into selected layers, showing that small parameter additions can effectively steer model behavior. Model-editing methods such as ROME~\citep{meng2022locating} and MEMIT~\citep{meng2022mass} further treat parameters as a writable memory substrate by inserting factual associations through localized or low-rank weight updates. However, these methods are less suited to online memory: their memory is usually fixed after training or updated through discrete editing steps, rather than evolving continuously with the sequence. Their write granularity is also less aligned with interaction history, which often unfolds at token-, message-, or segment-level resolution. As a result, parametric memory often acts as a persistent modification to model behavior, rather than a state-conditioned memory mechanism whose influence changes with the current history. \deltamem{} is close to LoRA in its low-rank interface, but differs fundamentally in that LoRA's low-rank update is static, whereas \deltamem{} generates low-rank attention corrections from a compact online state at runtime.

\paragraph{Hybrid Attention and Recurrent Memory.} 
The closest prior work can be divided into three groups. Hybrid sequence architectures, including NAtS-L~\citep{NASL}, NHA~\citep{NHA}, Hymba~\citep{Hymba}, combine softmax attention with linear attention, recurrent slots, or SSM/RNN branches to balance modeling quality and computational efficiency; these components constitute the model’s token-mixing backbone and maintain sequence-local rather than interaction-persistent states in their standard formulations. Infini-attention~\citep{infinitee} instead propagate compressed matrices or memory tokens across segments of a long input stream. Although Infini-attention also considers a delta-rule update, it fuses the retrieved memory with local attention. Titans~\citep{behrouz2024titans} and LongMem~\citep{wang2023augmenting} are more explicitly memory-oriented, but respectively update neural-memory parameters at test time or retrieve representations from an external memory bank through a residual side network. In contrast, $\delta$-Mem retains a frozen pretrained full-attention backbone, persists a fixed-size associative state across interactions, and uses the previous-state readout only to generate low-rank query- and output-side corrections. Only the lightweight $\delta$-Mem modules are trained with SFT, while inference updates the state without replaying the stored history. Thus, $\delta$-Mem is not merely another hybrid token mixer: its contribution lies in combining online persistence, gated delta writing, and state-conditioned attention steering. It provides an efficient pipeline to combine with current LLMs for local deployment and enterprise scenarios.

\section{Conclusion}
\label{sec:conclusion}

In this work, we introduced \deltamem{}, a lightweight memory mechanism that equips a frozen full-attention backbone with a compact and dynamically updated online state of associative memory. \deltamem{} compresses past information into a fixed-size online state and uses its readout to generate low-rank corrections to the backbone's attention components. This design allows memory to be maintained online and to directly participate in forward computation without full fine-tuning or replacing the backbone architecture.
Empirically, \deltamem{} improves performance on memory-heavy benchmarks while largely preserving the general capabilities of the frozen backbone. Notably, even with an extremely small $8\times8$ online state, the model can recover useful historical information after explicit context is removed, showing that effective memory does not require extending explicit context or heavy external retrieval modules. These results suggest that compact online states can serve as a scalable and efficient interface for test-time memory in frozen Transformer backbones.

\section*{Acknowledgment}
This research/project is supported by the National Research Foundation, Singapore under its AI
Singapore Programme (AISG Award No: AISG3-GV2023-010). This work is also supported by
the National Research Foundation, Singapore, under its National Large Language Models Funding
Initiative (AISG Award No: AISG-NMLP-2024-005), and NTU SUG project \#025628-00001: Posttraining to Improve Embodied AI Agents.
\bibliographystyle{plain}
\bibliography{references}

\newpage
\appendix
\section{Limitations}
\label{sec:limitation}
Although \deltamem{} provides online memory mechanism, we acknowledge two main areas for future refinement. First, as with most of memory mechanisms, the compact state inherently represents an approximation of the explicit context. While it significantly outperforms zero-history baselines, further aligning its representational capacity with exact full-context inference remains an important direction for future research. Second, the dynamic state update operations introduce marginal computational steps during decoding. However, since the GPU memory overhead is negligible, this theoretical impact on inference speed can be readily mitigated through system-level optimizations, which we defer to future work.

\section{Broader Impacts}
\label{sec:broad_impact}
$\delta$-Mem aims to improve the efficiency and effectiveness of memory-augmented language models by enabling historical information to be maintained in a compact online state rather than relying solely on explicit context expansion. This may benefit long-term assistants, interactive agents, and other applications that require persistent context under limited computational budgets. At the same time, stronger memory mechanisms may introduce risks related to privacy, unintended retention of sensitive information, accumulation of stale or incorrect memories, and potential user profiling. These risks are particularly relevant when memory is maintained across long interactions or deployed in personalized systems. Therefore, $\delta$-Mem should be used with appropriate safeguards, including explicit user consent, transparent memory controls, deletion mechanisms, and filtering or auditing procedures for sensitive information. We view $\delta$-Mem as a technical memory mechanism, and its deployment should be accompanied by application-specific safety and privacy protections.

\section{Inference Efficiency and Memory Use}
\label{appendix:efficiency_and_memory}
\begin{figure}[htbp]
    \centering

    \begin{subfigure}[t]{0.43\textwidth}
        \centering
        \includegraphics[width=\linewidth]{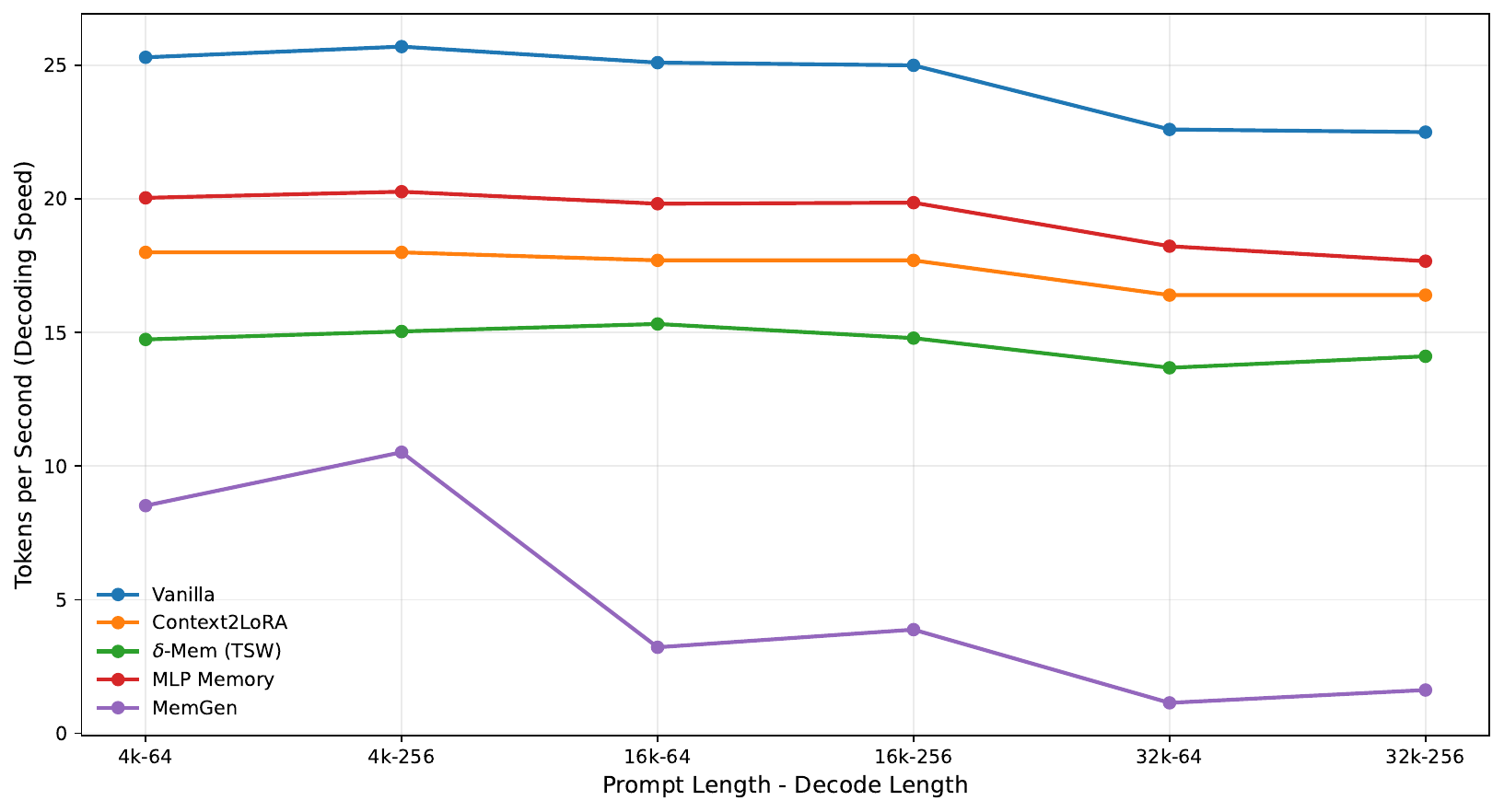}
        \caption{Decoding throughput.}
        \label{fig:decoding_tps}
    \end{subfigure}
    \hfill
    \begin{subfigure}[t]{0.48\textwidth}
        \centering
        \includegraphics[width=\linewidth]{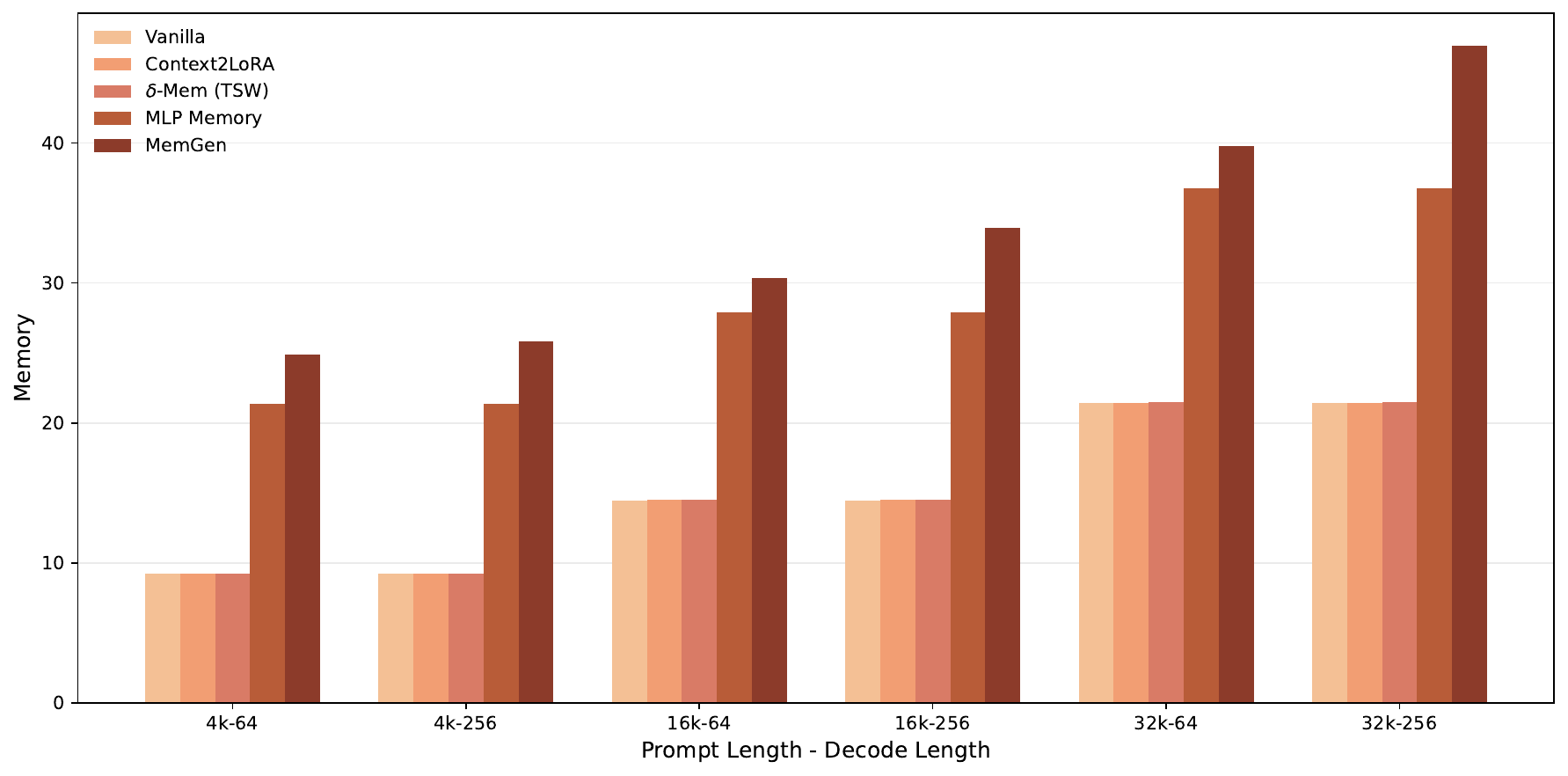}
        \caption{Memory usage.}
        \label{fig:memory_usage}
    \end{subfigure}

    \caption{Inference efficiency under different prompt and decode lengths.}
    \label{fig:inference_efficiency}
\end{figure}

We further compare the inference efficiency of different memory-augmented methods under varying prompt and decoding lengths, as shown in Figures~\ref{fig:memory_usage} and~\ref{fig:decoding_tps}. \deltamem{} achieves nearly the same GPU memory usage as Vanilla and Context2LoRA, indicating that its compact recurrent state introduces negligible memory overhead even when the prompt length increases to 32K. By contrast, MLP Memory and MemGen require substantially more memory, reflecting the cost of maintaining or generating larger auxiliary memory representations. In terms of decoding throughput, \deltamem{} is slower than Vanilla and Context2LoRA because each step involves reading from and updating the online state, but it remains considerably faster and more stable than MemGen across all tested settings. These results demonstrate that \deltamem{} improves long-context memory with a lightweight computational footprint, offering a practical balance between memory capability and inference efficiency.

\section{Implementation Details}
\label{appendix:setup}
\paragraph{Training Setup.}
All models are trained for one epoch on the shortest 2,219-sample split of QASPER~\citep{dasigi2021dataset}, whose maximum sequence length is 8,269 tokens. The maximum backbone training sequence length is set to 512, while the memory write budget is set to 8,192 tokens. Unless otherwise specified, \deltamem{} uses $r=8$ and $\alpha=16$, and is applied only to the query and output branches. The number of states in MSW is set to 4. Training is conducted on 8 $\times$ A800 GPUs with bfloat16 precision, DeepSpeed ZeRO-2~\citep{rasley2020deepspeed}, and fused AdamW. We use a peak learning rate of $2\times10^{-4}$ with cosine decay and a warmup ratio of 0.1. The per-device batch size is 1, with 4 gradient accumulation steps, resulting in an effective global batch size of 32. The random seed is fixed to 42.

\paragraph{Evaluation Details.}
We follow the official evaluation prompts and decoding settings for all benchmarks. Specifically, we report prompt-level strict accuracy for IFEval, Exact Match~(EM) and F1 for HotpotQA, accuracy for GPQA, and F1 for LoCoMo. For MemoryAgentBench, Table~\ref{tab:mab_details} summarizes the evaluation categories, datasets, and metrics. Each dataset is evaluated using its corresponding metric, and the final MemoryAgentBench score is computed as the sample-weighted average.
\begin{table}[htbp]
\centering
\caption{Evaluation categories, datasets, and metrics in MemoryAgentBench.}
\label{tab:mab_details}
\resizebox{0.6\linewidth}{!}{
\begin{tabular}{lll}
\toprule
\textbf{Category} & \textbf{Dataset} & \textbf{Metric} \\
\midrule
\multirow{4}{*}{\makecell[l]{Accurate\\Retrieval}}
& SH-Doc QA & \multirow{4}{*}{Accuracy} \\
& MH-Doc QA\\
& LongMemEval (S*) \\
& EventQA \\
\midrule
\multirow{6}{*}{\makecell[l]{Test-time\\Learning}}
& BANKING77 & \multirow{5}{*}{Accuracy} \\
& CLINC150  \\
& NLU \\
& TREC Coarse \\
& TREC Fine  \\
& Movie Recommendation & Recall@5 \\
\midrule
\multirow{2}{*}{\makecell[l]{Long Range\\Understanding}}
& $\infty$Bench-Sum & F1-Score \\
& Detective QA & Accuracy \\
\midrule
\multirow{2}{*}{\makecell[l]{Selective\\Forgetting}}
& FactConsolidation-SH & \multirow{2}{*}{Accuracy} \\
& FactConsolidation-MH  \\
\bottomrule
\end{tabular}
}
\end{table}

\section{Parameter Overhead}
\label{appendix:parameter}

\begin{figure}[htbp]
    \centering
    \includegraphics[width=0.7\linewidth]{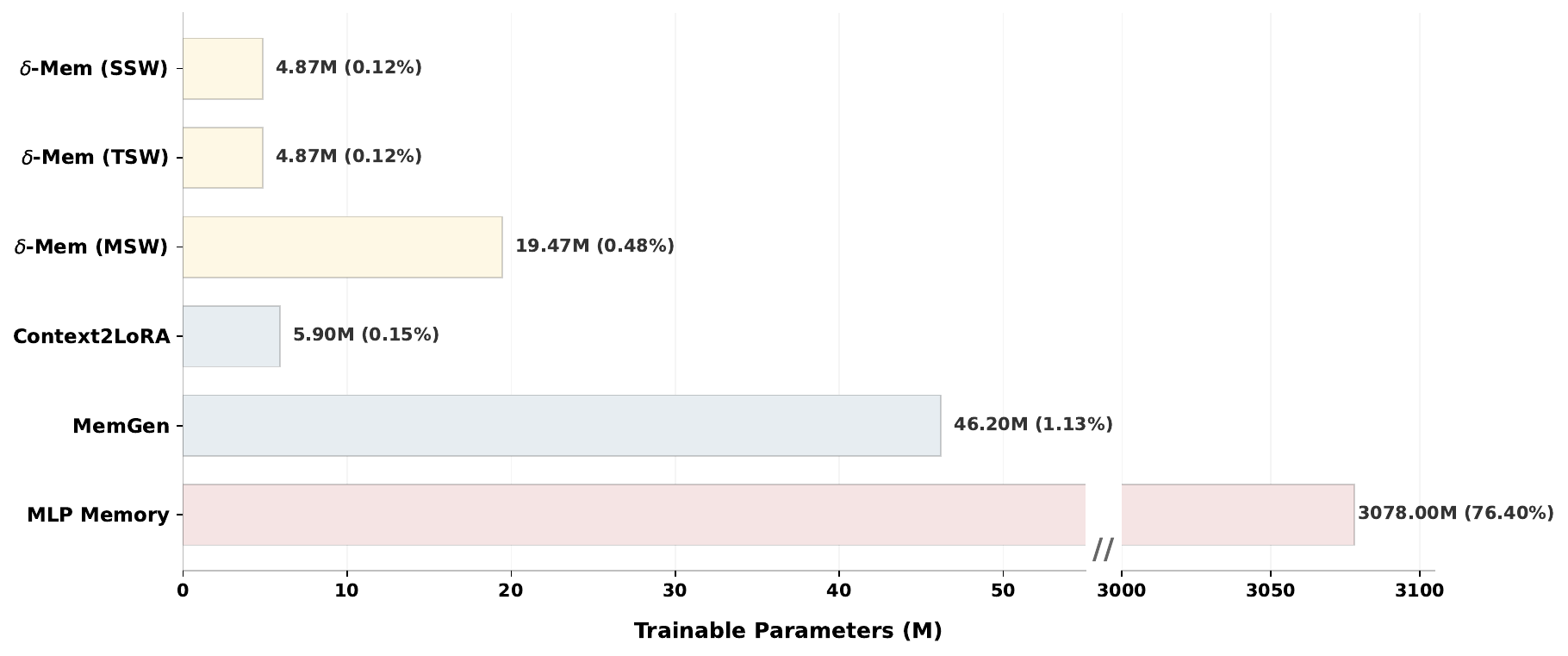}
    \caption{Trainable parameter comparison across memory-augmented methods.}
    \label{fig:params}
\end{figure}

We compare the trainable parameter overhead of \deltamem{} with representative memory-augmented baselines, as shown in Figure~\ref{fig:params}. \deltamem{} introduces only 4.87M trainable parameters for both SSW and TSW variants, accounting for merely 0.12\% of the backbone parameters. Even the MSW variant, which maintains multiple memory states, requires only 19.47M trainable parameters, corresponding to 0.48\% of the backbone. In contrast, MemGen uses 46.20M trainable parameters, while MLP Memory requires 3078.00M parameters, reaching 76.40\% of the backbone scale. These results show that \deltamem{} achieves online memory augmentation with substantially lower parameter overhead, making it a lightweight alternative to larger auxiliary-memory modules.


\end{document}